\documentclass[letterpaper, 10 pt, conference]{ieeeconf}  

\IEEEoverridecommandlockouts                              

\usepackage{amsmath} 
\usepackage{bbm}
\usepackage{dsfont}
\usepackage{xcolor}
\usepackage{amssymb}
\usepackage{graphicx}
\usepackage{mathtools}
\usepackage{amsmath}
\usepackage{gensymb}
\usepackage{float}
\usepackage{multirow}
\usepackage{makecell}
\usepackage{mathtools}
\usepackage{hyperref}
\hypersetup{
    colorlinks=true,
    linkcolor=blue,
    filecolor=magenta,      
    urlcolor=blue,
    pdftitle={Overleaf Example},
    pdfpagemode=FullScreen,
    }

\DeclarePairedDelimiter\floor{\lfloor}{\rfloor}
\usepackage{tabularx}
    \newcolumntype{L}{>{\raggedright\arraybackslash}X}
\usepackage[utf8]{inputenc}
\usepackage[english]{babel}

\usepackage{amsthm}

\usepackage{subcaption}

\newcommand{\no}{\noindent}

\newcommand\Tstrut{\rule{0pt}{2.6ex}}         
\title{\LARGE \bf  \ours: Legged Robot Navigation in Unstructured Outdoor Environments using Offline Reinforcement Learning
}

\author{Kasun Weerakoon$^{1}$, Adarsh Jagan Sathyamoorthy$^{1}$, Mohamed Elnoor$^{1}$, and Dinesh Manocha$^{2}$
\thanks{$^{1}$ Authors are with Dept. of Electrical and Computer Engineering, University of Maryland, College Park, MD, USA. {\tt\footnotesize kasunw@umd.edu, asathyam@umd.edu, melnoor@umd.edu}}
\thanks{$^{2}$ Author is with Dept. of Computer Science, University of Maryland, College Park, MD, USA. {\tt\footnotesize dm@cs.umd.edu}} \\
\small{Supplemental version including Tech Report, Code and Video at \url{http://gamma.umd.edu/vapor/}}}

\newcommand{\ours}{VAPOR} 

\begin{document}

\maketitle
\thispagestyle{empty}
\pagestyle{empty}

\begin{abstract}
We present \ours, a novel method for autonomous legged robot navigation in unstructured, densely vegetated outdoor environments using offline Reinforcement Learning (RL). Our method trains a novel RL policy using an actor-critic network and arbitrary data collected in real outdoor vegetation. Our policy uses height and intensity-based cost maps derived from 3D LiDAR point clouds, a goal cost map, and processed proprioception data as state inputs, and learns the physical and geometric properties of the surrounding obstacles such as height, density, and solidity/stiffness. The fully-trained policy's critic network is then used to evaluate the quality of dynamically feasible velocities generated from a novel context-aware planner. Our planner adapts the robot's velocity space based on the presence of entrapment inducing vegetation, and narrow passages in dense environments. We demonstrate our method's capabilities on a Spot robot in complex real-world outdoor scenes, including dense vegetation. We observe that \ours's actions improve success rates by up to 40\%, decrease the average current consumption by up to 2.9\%, and decrease the normalized trajectory length by up to 11.2\% compared to existing end-to-end offline RL and other outdoor navigation methods. 
Code implementation is available \href{https://github.com/kasunweerkoon/VAPOR}{here}.



\end{abstract}

\section{Introduction} \label{sec:intro}

Autonomous robot navigation in complex outdoor scenes is an essential capability for many applications, including precision agriculture \cite{naik2016precision}, search and rescue operations in forested environments \cite{karma2015use}, reconnaissance\cite{li2019fire}, etc. There are two major challenges in navigating such scenarios. Firstly, the robot must perceive and differentiate non-solid/pliable obstacles (e.g. tall grass), from solid/non-pliable obstacles (e.g. trees) \cite{sathyamoorthy2023vern}. Pliable obstacles can be safely traversed through, whereas non-pliable obstacles must be avoided. Secondly, apart from avoiding collisions, the robot also faces challenges such as narrow passages, and scenarios where the vegetation could wrap/attach onto the robot and entrap it. The robot’s navigation must be capable of handling such adverse situations.

To address the perceptional challenges in outdoors, methods based on image classification \cite{sathyamoorthy2023vern}, semantic segmentation \cite{9810192}, and anomaly detection \cite{wellhausen2020safe} using supervised learning have been employed. However, such works require extensive manual annotation and labeling to identify traversable terrain during training. Such models also may not align with the actual traversability capabilities of different robots due to varying dynamic constraints. This restricts the robot's navigation and could lead to highly conservative, meandering trajectories \cite{sathyamoorthy2022terrapn}, or freezing behaviors \cite{9099106}. Imitation learning techniques have also been proposed for outdoor navigation, but the resulting models may not generalize well\cite{hussein2018deep}.

\begin{figure}[t]
      \centering
      \includegraphics[width=0.84\columnwidth,height=5cm]{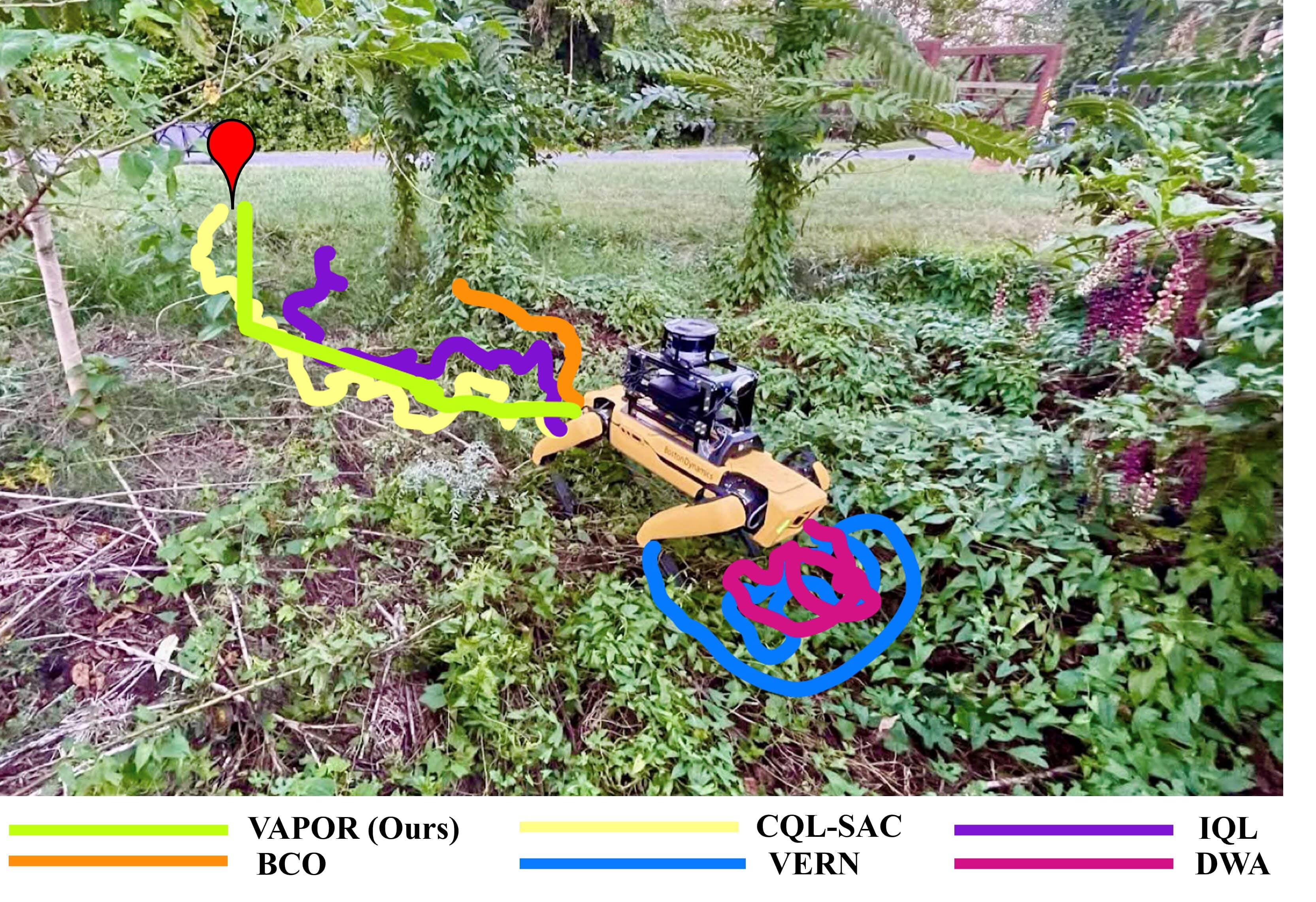}
      \caption {\small{Legged robot trajectories generated when navigating through complex outdoor vegetation using \ours, end-to-end CQL-SAC \cite{kumar2020CQL}, IQL \cite{kostrikov2021IQL}, BCO \cite{torabi2018BCO}, VERN \cite{sathyamoorthy2023vern}, and DWA \cite{fox1997dynamic}. \ours{} identifies the entrapment in vines and uses holonomic actions to minimize the instability and current consumption instead of excessive angular actions taken by the other methods which results in further entrapment. Hence, in this scenario, \ours{} moves backward with minimal angular motion to reduce entanglement with vines.}}
      \label{fig:cover-image}
      \vspace{-25pt}
\end{figure}

Conversely, outdoor navigation methods based on online reinforcement learning (RL) \cite{weerakoon2022terp} do not require human labeling since they are trained using a robot's active interactions with a simulated environment. Nevertheless, such models exhibit severe performance degradation during real-world deployment due to sim-to-real transfer issues \cite{liang2021crowd}. Training complex models using RL requires high-fidelity simulations, which may not be available, especially for complex scenarios. To alleviate such shortcomings, offline-RL \cite{levine2020offline} methods have been proposed, where a model is trained using data collected in real-world environments, reducing the sim-to-real transfer issues.

However, none of the existing methods for outdoor navigation have accounted for the constraints that complex real-world scenes imposes on the robot's velocities. For instance, while traversing through tall grass and bushes, angular motions could cause the vegetation to easily wrap around the robot restricting its motion. Furthermore, in cases with solid obstacles create narrow passages, the robot would have to rotate/reorient itself to maneuver the narrow free space. Therefore, the robot's executable actions must be adapted based on the environment.     

\textbf{Main contributions:} To address these challenges, we propose \ours, an offline RL-based trajectory evaluation model combined with a context-aware planner designed to generate dynamically feasible velocities to operate a legged robot in challenging outdoor scenes. \ours's offline RL formulation allows it to be trained using data collected in the real world  \cite{shah2022offline} that is automatically compiled for training, alleviating sim-to-real transfer issues. The novel components of our work are:

\begin{itemize}

\item We propose a novel offline RL-based actor-critic network to learn a Q-function to evaluate a legged robot's candidate actions and velocities in terms of their ability to reach the goal, avoid solid, non-pliable vegetation and other desirable behaviors. The network consists of spatial and channel attention layers to learn the spatial correlations in the input observation space. Our model is trained using real-world data collected in dense environments that are automatically compiled into states and actions between randomly chosen start and goal states. This alleviates the sim-to-real transfer issues prevalent in existing RL methods. This results in
an improvement up to 40\% in terms of success rate 

\item A novel observation space to sense dense vegetation consisting of robot-centric height and intensity cost maps obtained by processing lidar point clouds, a goal map indicating the distance and direction to the goal, and proprioceptive signals from the legged robot's joints to indicate its stability. The height and intensity maps accurately represent the height and solidity or inversely, the pliability of the surrounding vegetation. The goal map and proprioception aid with spatially correlating the vegetation's properties with the robot's intended motion direction and stability during training. 

\item A novel context-aware motion planner that switches between (1). a holonomic velocity space to minimize the risk of entrapment in vegetation assessed from proprioceptive signals, and (2). a non-holonomic velocity space to navigate narrow passages between solid, non-pliable vegetation. Further, it generates dynamically feasible, smooth candidate actions/velocities to be assessed by \ours's Q-function. \ours is evaluated on a real Boston Dynamics Spot robot in unstructured outdoor scenes.  
\end{itemize}

\section{Related Work} \label{sec:related_work}
In this section, we discuss the existing literature on vegetation perception in outdoor environments, and offline RL methods used for navigation. Finally, we discuss the existing holonomic planning methods.

\subsection{Outdoor Vegetation Perception}

Navigating robots in outdoor environments, particularly through vegetation, is a challenging task that has received increasing attention in recent years \cite{sathyamoorthy2023vern,iqbal2020simulation,9345970}. Existing approaches tackle this issue using various sensory modalities and learning techniques. For instance, \cite{li2023seeing} adopts a self-supervised approach to estimate the support surface in vegetation employing 3D point clouds and RGB images. Despite their promising results, the system requires manual labeling, which could be time-consuming and less scalable for real-world deployments. In \cite{sathyamoorthy2023vern}, the authors use RGB images and 2D Lidar to create traversability cost maps in dense vegetation environments. Stone et al. \cite{stone2019vegetation} use an infrared (IR) sensor and an RGB camera for vegetation detection. While effective in certain conditions, these camera-based methods are often vulnerable to environmental factors such as changing lighting and motion blur, thereby limiting their robustness. Iberraken et al. \cite{iberraken2022autonomous} demonstrate the use of a 2D LiDAR to navigate through structured vineyard fields.

Some recent works have shifted their focus from external sensors to proprioceptive modalities to perceive vegetation \cite{elnoor2023pronav,jian2023path}. While proprioception offers reliable feedback about the robot's internal state, it inherently lacks the capability for look-ahead predictions before traversing a given terrain, especially in the absence of exteroceptive sensors. Our work combines exteroception (3D point clouds) with proprioception for robust and efficient navigation through outdoor vegetation. 

\subsection{Offline RL based Robot Navigation}


Reinforcement Learning (RL) has been fundamental to robot navigation \cite{weerakoon2022terp,faust2018prm,dugas2021navrep,patel2021dwa}, providing methods for autonomous decision-making based on interaction with the environment. However, traditional online RL often falls short in situations where real-time data collection is either impractical or the lack of realistic simulation which increases the sim-to-real gap \cite{zhang2021reinforcement} e.g., navigating through dense vegetation or hazardous or complex terrains. On the other hand, offline RL has emerged as a promising alternative, designed to optimize policies based on pre-collected datasets. Among the foundational works in offline RL is the study by Levine et al. \cite{levine2020offline}, which outlines the key methodologies and challenges. Methods based on Imitation Learning (IL) have also leveraged collected data \cite{codevilla2018end,silver2010applied}. However, IL is often restricted by the limitations of the human operator (expert) who collected the data, meaning it cannot generally surpass the operator's performance. In contrast, offline RL aims to optimize the behavioral policy based on the dataset, offering the potential for more generalized and sometimes superior strategies \cite{kostrikov2021IQL,li2022hierarchical,shah2022offline}. Kostrikov et al. \cite{kostrikov2021IQL} introduce Implicit Q-Learning (IQL) which implicitly estimates the value function without querying the Q function of unseen actions. While IQL shows promise, it struggles in tasks with long planning horizons. Shah et al. \cite{shah2022offline} mitigate this limitation by combining IQL with topological graphs. Nevertheless, their method primarily relies on RGB images which is susceptible to lighting changes, motion blur, etc.

On the other hand, \cite{kumar2020CQL} developed Conservative Q-Learning (CQL) to enhance the robustness of the learned policy. This method lower-bounds the true value of its learned Q function. Following its superior performance with complex data distribution, we extend this method by employing data from a 3D LiDAR and a legged robot's joint encoders.


\subsection{Holonomic Planning}
Traditional robotic planning often focuses on non-holonomic robotic planners, largely because many robots including wheeled robots, inherently possess non-holonomic constraints \cite{yang2014spline,khan2017complete,chae2020robust,eshtehardian2023continuous}. Conversely, robots with higher degrees of freedom (e.g., legged or manipulator robots) can benefit from holonomic planning methods \cite{holmberg2000development, alireza2021experimental}. However, such planner lack the ability to adapt the robot's velocity space based on environmental constraints, especially in dense vegetation.

\section{ \ours: Vegetation-Aware Planning using Offline Reinforcement Learning} \label{sec:our-method}

\begin{figure}[t]
      \centering
      \includegraphics[width=\columnwidth,height=4.5cm]{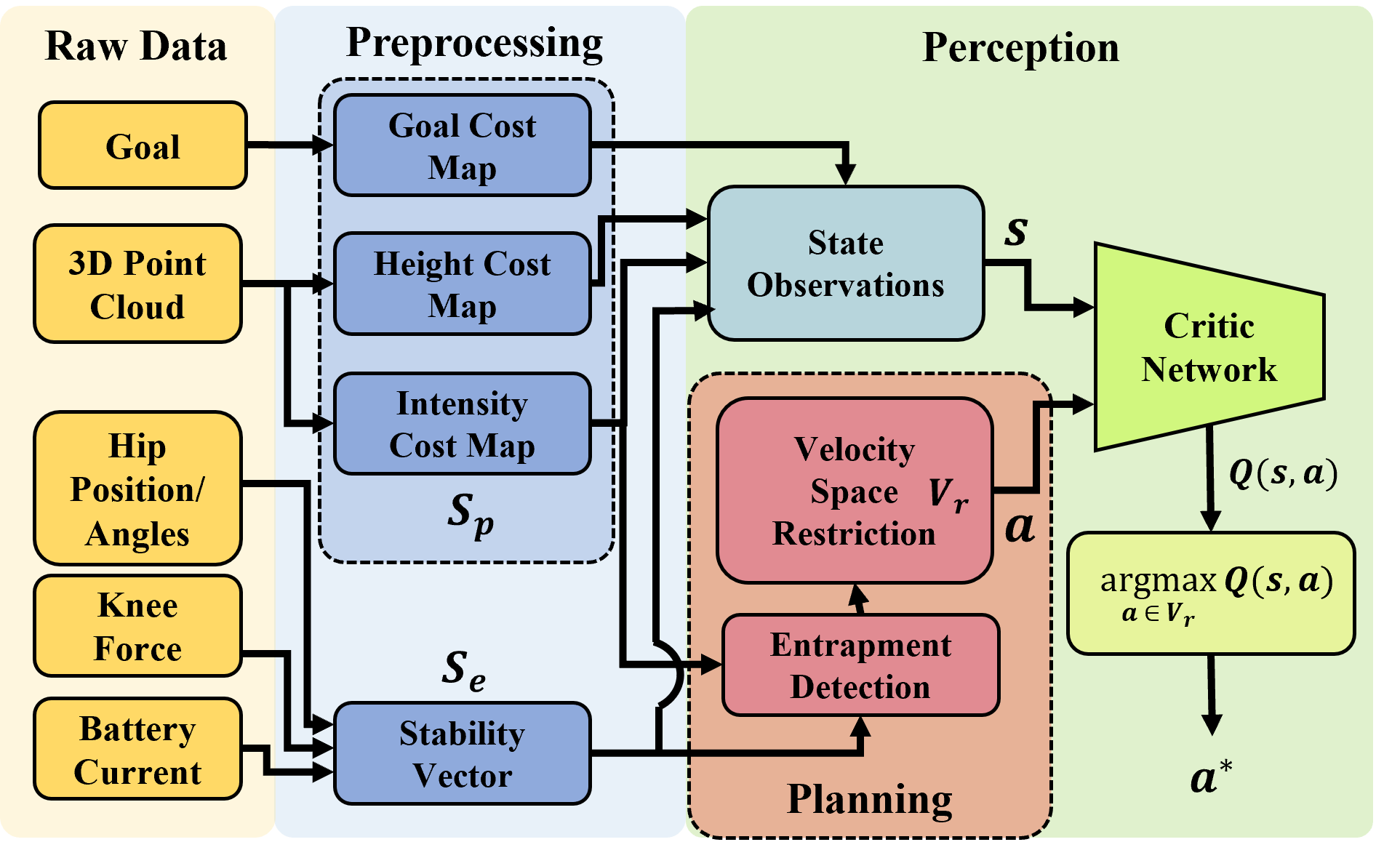}
      \caption {\small{Overall system architecture of \ours{} which uses a height and intensity cost map generated from 3D lidar, a goal cost map, and proprioception data from the robot as state inputs to train an Actor-Critic offline RL policy. Then, the fully trained critic network is used to evaluate the dynamically feasible actions generated by a planner. The planner uses instability detection using proprioception and intensity map to switch between a holonomic and non-holonomic action space to reduce the risk of entrapment.  }}
      \label{fig:System_arch}
      \vspace{-20pt}
\end{figure}

\subsection{Preliminaries}
We mathematically formulate our navigation problem as a Markov Decision Process (MDP) with continuous states and actions. Our MDP can be defined as $\mathcal{M}:= \{ \mathcal{S}, \mathcal{A},  \mathbb{P}, r, \gamma\}$, where $\mathcal{S}, \mathcal{A}$ denote state and action spaces, $\mathbb{P}(s^{\prime}|s,a)$ represents the state transition dynamics between current state $s$, action $a$, and next state $s^{\prime}$. $r(s,a)$ is the reward function, and $\gamma \in (0,1)$ denotes the discount factor. The objective of RL is to learn a policy $\pi_{\theta}(a|s)$ parameterized by $\theta$ that maximizes the discounted cumulative reward return.

Offline RL particularly aims to learn policies from existing data sets instead of explicitly interacting with the environment. Hence, for a dataset $\mathcal{D}=\{ (s_j,a_j,r_j,s_j^{\prime})| s_j,s_j^{\prime} \in \mathcal{S}; a_j \in \mathcal{A} ; j=1,2,..,N\}$, offline RL algorithms attempt to learn a policy $\pi_{\theta}(a|s)$ that maximizes the discounted reward return $R_t = \sum_{k=t}^{T} \gamma^{(k-t)}r_k(s_k,a_k)$ at time step $t$. However, leveraging the standard RL algorithms for offline RL leads to poor performance due to overfitting and distributional shifts \cite{levine2020offline}.  In particular, the existing value-based off-policy RL methods such as Q learning typically overestimate the value function predictions for unseen outcomes, which results in erroneous and overly optimistic estimations \cite{prudencio2023survey_offrl}. To mitigate this issue,  Conservative Q Learning (CQL) \cite{kumar2020CQL} regularizes the Q-values during training to learn conservative and lower-bound estimates of the value function. Hence, in this work, we incorporate CQL with Soft Actor-Critic (SAC) \cite{haarnoja2018soft} as our base offline RL algorithm. 

Hereafter, we use $j, k$ as indices. Vectors are represented in bold, lower case letters. All positions, velocities, and forces are represented w.r.t a rigid frame attached to the robot $R$ (indicated in superscript) or relative to a cost map. The robot frame's $x, y, z$ directions points forward, leftward, and upward respectively. 

\subsection{Dataset Generation}
 Our raw training data is collected by teleoperating a legged robot equipped with a $360^{\circ}$ 3D LiDAR, and joint encoders for $\sim 4$ hours. We collect raw 3D point clouds, robot's odometry, joint positions and velocities on the legs, and joint actuator current as the robot moves in random trajectories in vegetation including grass, bushes, and trees with varying density. Hence, the raw data set does not have any goal-conditioning or goal-reaching policy.

To create goal-conditioned data set $\mathcal{D}$ with a series of $\{s_j, a_j, r_j, s^{\prime}_j\}$, we consider random trajectory segments from the raw dataset, i.e., we select a random state as the initial position and a future sample in the same raw trajectory as the goal. This subsequent goal sample is selected such that it is $\sim 8-20$ meters away from the robot's initial position, and our processed dataset $\mathcal{D}= \{ (s_j,a_j,r_j,s_j^{\prime}) \, | \, j=1,2,..,N\}$ is obtained. We explain the details of the state observations, actions, and reward formulation in the sub-sections below.


\subsection{State Observations from Multi-sensor Data}

Our state observations $ s \in \mathcal{S}$ are obtained by pre-processing the raw sensory data collected from both the exteroceptive (point clouds) and proprioceptive (joint positions, forces) data from the robot. We denote the entire point cloud as $\mathbf{P}$ reflected point's 3D location relative to the robot and intensity as $\mathbf{p}_{j} = \{x_j, y_j, z_j, i_j\} | x_j, y_j, z_j \in \mathbb{R}; i_j \in [0, i_{m}] \}$. Proprioceptive sensing is obtained from the robot's joint positions $h_1^{x/y}, h_2^{x/y}, h_3^{x/y}, h_4^{x/y}$, force feedback $f_{1}, f_{2}, f_{3}, f_{4}$, and the battery's current consumption $I_b$.


We preprocess the aforementioned sensory data to generate two types of state observations: 1.) $S_{e}$: A set of robot-centric cost maps that reflect the solidity and height of the surrounding objects, and distance to the goal using exteroceptive sensors; 2.)  $S_{p}$: A vector that quantifies the robot's stability using proprioception. Hence, our final state observations $ s = [S_{e},S_{p}] \in \mathcal{S}$.

\begin{figure}[t]
      \centering
      \includegraphics[width=0.9\columnwidth,height=2.6cm]{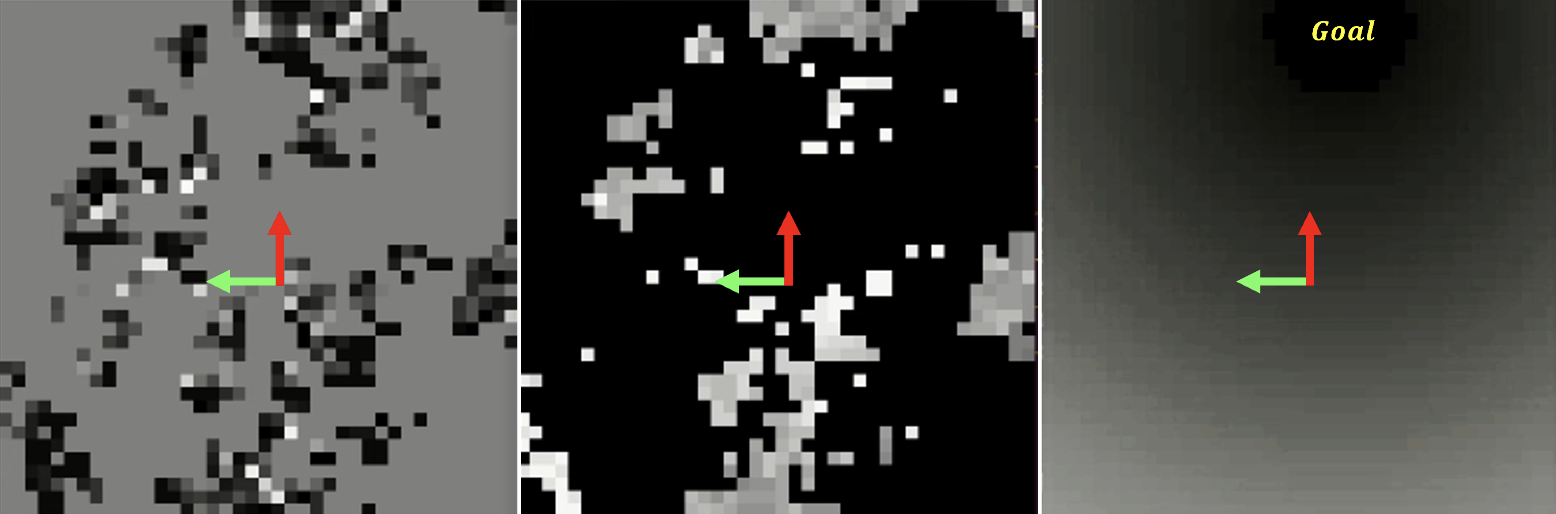}
      \caption {\small{Robot-centric cost map state observations from exteroceptive sensing from Scenario 3 in Fig. \ref{fig:comparison_trajs}: \textbf{[Left]} Point cloud-based intensity cost map $\mathcal{C}_i$ that indicates the density of the surrounding objects using lidar reflectance; \textbf{[Center]} Height cost map $\mathcal{C}_h$ that represents the maximum height of the objects derived from the point cloud; \textbf{[Right]} Goal cost map $\mathcal{C}_g$ that indicates the distance to the goal from the robot's neighborhood. Light colors indicate higher costs and dark areas represent lower costs. }}
      \label{fig:cost_maps}
      \vspace{-20pt}
\end{figure}

\subsubsection{Layered Cost Maps from Exteroception}
Navigation in outdoor vegetation requires sensing the height and solidity of the vegetation in the robot's vicinity. Moreover, spatial information of the goal location is necessary to perform successful goal-reaching tasks. Hence, we propose three robot-centric 2D cost maps, intensity cost map $(\mathcal{C}_i)$, height cost map $(\mathcal{C}_h)$,  and goal cost map $(\mathcal{C}_g)$, to represent the solidity and height of the surrounding objects/vegetation, and distance and direction to the goal respectively. 

All three cost maps $\mathcal{C}_i, \mathcal{C}_h$ and, $ \mathcal{C}_g $ are $n \times n$ matrices with the robot positioned at the center $(n/2, n/2)$ as depicted in Fig. \ref{fig:cost_maps}. Each element in each cost map satisfies $ \mathcal{C}_{i, h, g}(l,m) \in [0,100] \quad \forall \, l,m = 0,1,.., n-1$. A grid $(l, m)$ in a cost map is related to a grid $grid^R_{l, m}$ of physical $(x, y)$ locations relative to the robot as,

\vspace{-15pt}
\begin{equation}
\begin{split}
    (x, y) &\in grid^R_{l, m} \\
    grid^R_{l, m} &= [[x_{l, m}, x_{l, m} + \beta], [y_{l, m}, y_{l, m} + \beta]] \\
    x_{l, m} &= \floor*{(l - \frac{n}{2}) \cdot \beta} \,\,\, \text{and} \,\,\, y_{l, m} = \floor*{(m - \frac{n}{2}) \cdot \beta},
\end{split}
\end{equation}

where $\beta$ is the side length of a square-shaped grid $grid^R_{l, m}$ in meters.


\no \textbf{Intensity Cost Map $(\mathcal{C}_i)$: } 
We employ the point cloud intensity values \cite{di2021visual,barfoot2016into,weerakoon2022cartographerglass}, to construct an intensity cost map $\mathcal{C}_i$. The LiDAR's reflectance power (i.e., intensity) is directly proportional to the solidity of the corresponding objects. Hence, we observe that grass, bushes, and trees result in distinct intensities (see Fig. \ref{fig:cost_maps}a). We calculate elements of $\mathcal{C}_i$ as,
\vspace{-5pt}
\begin{equation}
    \mathcal{C}_i(l,m) = \frac{\sum_{x_j}\sum_{y_j} i_j}{\beta^2} \,\,\, \forall \mathbf{p}_j \in \mathbf{P} \,\, \text{and} \,\, x_j, y_j \in grid^{R}_{l, m}.
\end{equation}

\no \textbf{Height Cost Map $(\mathcal{C}_h)$: } 
We generate $\mathcal{C}_h$ to represent the maximum heights of the objects in each grid location $grid^{R}_{l, m}$. To this end, element $(l,m)$ of $\mathcal{C}_h$ is obtained by,
\begin{equation}
    \mathcal{C}_h(l,m) = max(z_j) \,\,\,  \forall \mathbf{p}_j \in \mathbf{P} \,\, \text{and} \,\, x_j, y_j \in grid^{R}_{l, m},
\end{equation}

\no where higher values in $\mathcal{C}_h(l,m)$ indicate taller objects.

\no \textbf{Goal Cost Map $(\mathcal{C}_g)$: }
Each location $(l, m)$ in the goal cost map represents $grid^{R}_{l, m}$'s distance to the goal $(x^R_g, y^R_g)$. Its value is calculated as, 

\begin{equation}
    \mathcal{C}_g(l,m) = \frac{\alpha_g .\Big(\sqrt{(x_g^R - x_{l, m})^2 + (y_g^R - y_{l, m})^2} \Big)}{d_{tot}},
\end{equation}

\no where $d_{tot}$ is the total distance to the goal from the robot's starting position and $\alpha_g \in \mathbb{R}$ is a tunable weight parameter.

Finally, we obtain our state observation from the exteroception $S_{e}$ by concatenating the derived cost maps. Hence, $S_{e} = \{\mathcal{C}_i, \mathcal{C}_h,  \mathcal{C}_g\} $ of shape $n \times n \times 3$.

\subsubsection{Stability Observation from Proprioception}

To estimate the robot's stability in vegetation, we incorporate data acquired from the robot's joint positions, forces, and battery current for proprioceptive sensing. To this end, we process the raw proprioceptive data $H_{prop}= h_1^{x/y}, h_2^{x/y}, h_3^{x/y}, h_4^{x/y}, f_{1/2/3/4}, I_b]$, as performed in \cite{elnoor2023pronav}. Principal Component Analysis (PCA) is then applied to the processed data to reduce its dimensions to two primary axes. Subsequently, we extract the variances ($\sigma^2_{PC1}$ and $\sigma^2_{PC2}$) of the dimension reduced data along the principal components, and define our resulting proprioceptive state observation vector as, 
 $S_{p} =[\sigma^2_{PC1}, \sigma^2_{PC2}], \sigma^2_{PC1}, \sigma^2_{PC2} \in \mathbb{R}^+$. We observe that highly stable terrains such as asphalt lead to lower variances, and unstable terrain leading to higher values.

Lastly, we derive our final state observation as $s = [S_{e}, S_{p}]$ by combining both exteroceptive and proprioceptive state observations.

\subsection{Offline Reinforcement Learning Using CQL-SAC}


Our network architecture is based on CQL-SAC \cite{kumar2020CQL} that incorporates two critic networks and an actor-network. The policy actor-network (i.e.,  $\pi_{\theta}(a|s)$) estimates the parameters $\theta$ of the policy distribution, which provides the conditional probability of taking action $a$ given the state observation $s$. In our context, this policy distribution is Gaussian parameterized by the mean $\mu_{\theta}$ and standard deviation $\sigma_{\theta}$. Further, the two critic networks are Q networks \big(i.e., $Q_{1}(s,a;\pi_{\theta}),Q_{2}(s,a;\pi_{\theta})  $\big)  that uses state-action pairs $s, a$ as inputs to estimate the expectation of the value function.  We design the actor and critic networks as follows.

\subsubsection{Actor and Critic Networks}
\begin{figure}[t]
      \centering
      \includegraphics[width=\columnwidth,height=4cm]{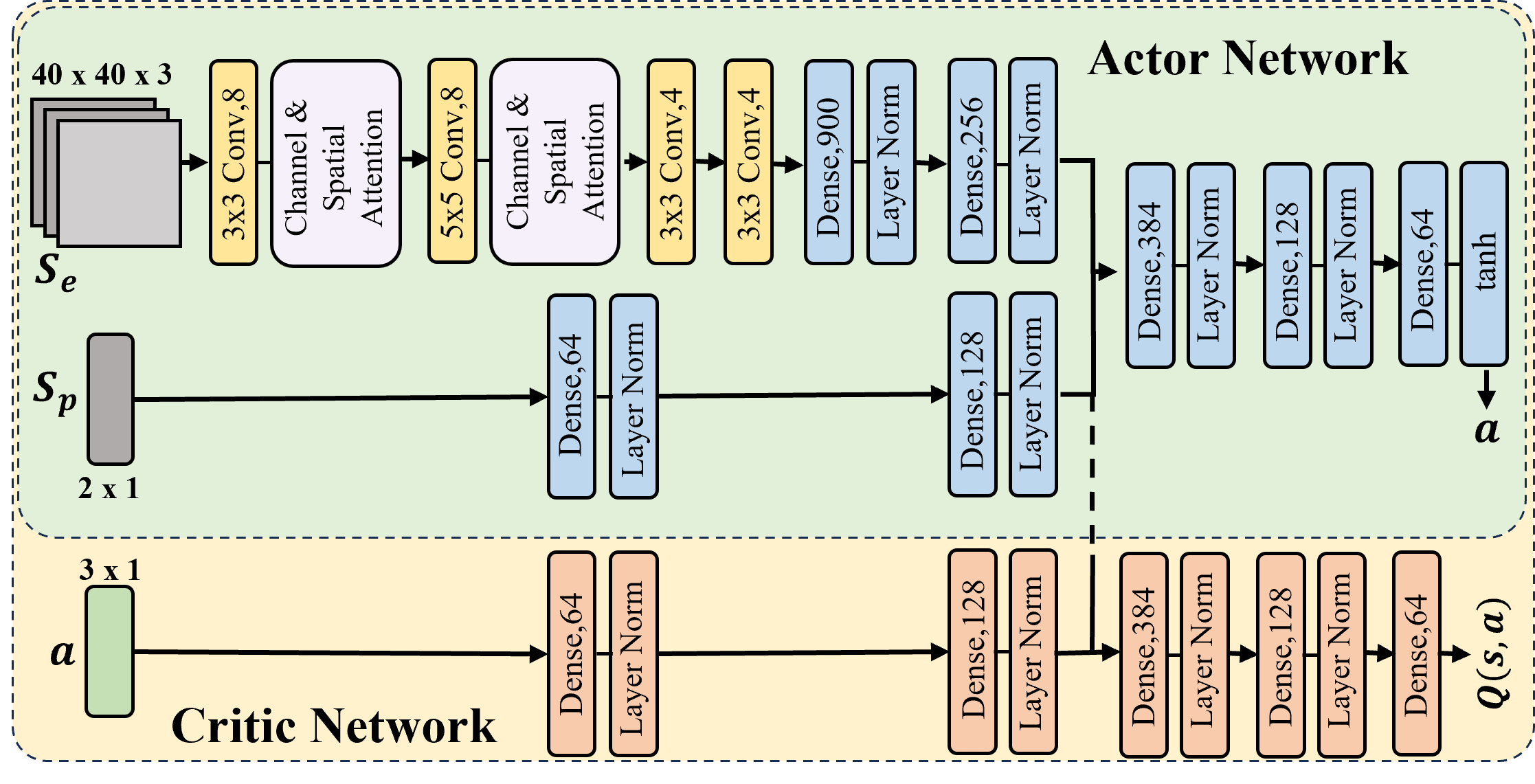}
      \caption {\small{Actor and Critic network architectures of our method. We incorporate two separate branches to process exteroception $S_e$, and proprioception $S_p$ observations. We use spatial and channel attention to encode correlation among the layered cost maps in $S_e$.}}
      \label{fig:network-arch}
      \vspace{-18pt}
\end{figure}

Our actor and critic network architecture with layer dimensions is presented in Fig. \ref{fig:network-arch}. In both networks, we use two separate network branches to process the exteroceptive $S_{e}$ and proprioceptive $S_{p}$ observations in our input state $s$. We highlight the use of spatial and channel attention networks in the exteroception branch. Spatial attention blocks encode spatial neighborhood properties in individual cost maps and channel attention helps learn the correlations between the features between the cost maps. The outputs from the two branches are concatenated and processed using several linear layers to obtain the end-to-end action outputs.


Since the critic networks take both the action and state inputs, we use an additional branch to process the action by passing two linear layers through before concatenating with the state observation branches.  All the hidden layers in the network are followed by $ReLU$ activation.

\subsubsection{Reward Functions}
The reward function is formulated to obtain robot actions that lead to desired navigation behavior. In this work, we are primarily interested in three navigation behaviors: 1) Goal reaching; 2) Avoiding dense/solid objects while navigating through pliable vegetation; and 3) Minimizing the overall energy consumption. We introduce three reward terms $r_{goal},r_{veg}$, and $r_{energy}$ to achieve the aforementioned behaviors. Hence, the total reward $r_{tot}$ obtained by the robot for a given sample is calculated as,  

\begin{equation}
    r_{tot} = \beta_1 r_{goal} + \beta_2 r_{veg}+\beta_1 r_{energy},
\end{equation}

where $\beta_1, \beta_2, \beta_3$ are tunable parameters to weigh the reward terms. We design $r_{goal}$ based on the robot's current distance $d_g$ to the goal to encourage moving towards the goal. Hence,

\begin{equation}
    r_{goal} = \frac{\lambda_{1} d_{tot}}{d_g}\mathds{1}_{\{d_g > d_{th}\}} + \lambda_{2}\mathds{1}_{\{d_g \leq d_{th}\}}, 
\end{equation}

 where $\lambda_{1},\lambda_{2} \in \mathbb{R}$ are adjustable parameters, $\mathds{1}$ is an indicator function, and $d_{th}$ is the goal reaching threshold. 
 
The vegetation reward $r_{veg}$ is a penalty for actions that navigate the robot in dense vegetation nearby (i.e, higher the density, lower the reward). To this end, we consider three circular neighborhoods with radii $0.5, 1.5$ and $2.5$ meters centered at the robot. Let, $A_{1}, A_2$ and $A_3$  denote the sets of grids corresponding to these neighborhoods in the intensity cost map $\mathcal{C}_i$. Then, $r_{veg}$ is calculated as,

\begin{equation}
    r_{veg} = -   \sum_{k=1,2,3} \bigg(\frac{\eta_{k}}{|A_k|}\sum_{l,m \in A_k}\mathcal{C}_i(l,m)\bigg),
\end{equation}

where the tunable parameters are set such that $\eta_{1} > \eta_{2} > \eta_{3} \in \mathbb{R}$ to ensure higher penalties for the dense vegetation in the robot's nearby vicinity. $|A_k|$ denotes cardinality of the set $A_k$.

We incorporate $r_{energy}$ to penalize actions consuming high amounts of energy (proportional to the current $I_b$) during navigation. We calculate $r_{energy}$ as,
 
\begin{equation}
    r_{energy} = - \epsilon I_b,
\end{equation}

where $\epsilon \in \mathbb{R}^+$ is a weight parameter.

\subsubsection{Critic Networks for State-action Evaluation} 
Even though we train an end-to-end navigation policy using CQL-SAC on our data set $\mathcal{D}$, we do not use the actions from the trained policy $\pi_{\theta}(a|s)$ in the actor network for navigation. Instead, we leverage the Q-function $Q(s,a)$ learned by a critic network to evaluate the quality of the set of actions generated by a context-aware planner. Intuitively, $Q(s,a)$ indicates how well the action leads to desirable behaviors imposed by the reward function. Since CQL-SAC includes two critic networks and learned Q-functions ($Q_{1}(s,a;\pi_{\theta})$ and $Q_{2}(s,a;\pi_{\theta})$), we choose the critic network with the lowest training loss. We refer to its Q-function as $Q_{min}(s,a;\pi_{\theta})$ from here on.

\subsection{Context-Aware Planning} \label{sec:holonomic-planning}
To generate dynamically feasible candidate actions to be evaluated using $Q_{min}(s,a;\pi_{\theta})$ , we formulate a novel context-aware planner. An action $a \in \mathcal{A}$ for our robot can be denoted as $a = (v_x, v_y, \omega_z)$. The planner uses a 3-dimensional velocity space ($V_s \subset \mathcal{A}$) defined as $V_s = \{(v_x, v_y, \omega_z) | -v_{max} \le v_x, v_y \le v_{max},  -\omega_{max} \le \omega \le \omega_{max}\}$. Here, $v_x$ and $v_y$ denote the linear velocities along the robot's x and y directions respectively, and $\omega_z$ represents the angular velocity about the vertical z-axis.   $v_{max}$ and $\omega_{max}$ are the maximum linear and angular velocity limits. Additionally, the planner uses the set of reachable/dynamically feasible velocities from the current velocities within an interval $\Delta t$ based on acceleration limits as $V_{r} = \{[v_x - \Dot{v}_{max}\Delta t, v_x + \Dot{v}_{max}\Delta t], [v_y - \Dot{v}_{max}\Delta t, v_y + \Dot{v}_{max}\Delta t], [\omega_z - \Dot{\omega}_{max}\Delta t, \omega_z + \Dot{\omega}_{max}\Delta t]\}$. Here, $\Dot{v}_{max}$, and $\Dot{\omega}_{max}$ are the robot's maximum linear and angular acceleration limits. 


The risk of entrapment in dense vegetation a robot faces is exacerbated when the robot performs angular motions because it aids the vegetation in helically twirling on to its legs (intuitively similar to rotating a fork on spaghetti). Therefore, in such scenarios, the robot's angular motion must be restricted. On the other hand, in scenarios with narrow passages, the rectangularly shaped robot must be capable of performing angular motions to traverse through. Such behaviors are also desirable when the robot is equipped with a sensor with a limited FOV that needs to be pointed in a specific direction. To accommodate both scenarios, we restrict $V_s$ based on the following condition:

\begin{equation}
\begin{split}
    \text{C}: \sqrt{\sigma^2_1 + \sigma^2_2} &> \Gamma, \,\, \mathcal{C}_i(l, m) \in [0.5 i_{m}, 0.75 i_{m}]
    \forall \,\,\, l, m \in A_2. \\
    V_s &= \begin{cases}
        \{(v_x, v_y, 0)\}, \,\,\, \text{if C is True},\\
        \{(v_x, 0, \omega_z)\}, \,\,\, \text{Otherwise},
    \end{cases}
\end{split} 
\end{equation}



\no where $v_x, v_y \in [-v_{max}, v_{max}]$, and $\omega \in [-\omega_{max}, \omega_{max}]$. The corresponding $V_r$ is calculated from the restricted $V_s$ by omitting either $v_y$ or $\omega_z$ based on the environment. The best action $a^*$ for the robot to execute given the current state $s$ can then be found as,
\vspace{-5pt}
\begin{equation}
    a^* = \underset{a_k \in V_r}{\operatorname{argmax}}(Q_{min}(s, a_k)). 
\end{equation}


\section{Results and Analysis}

\begin{figure*}[t]
    \centering
    \includegraphics[width=1.7\columnwidth,height=3cm]{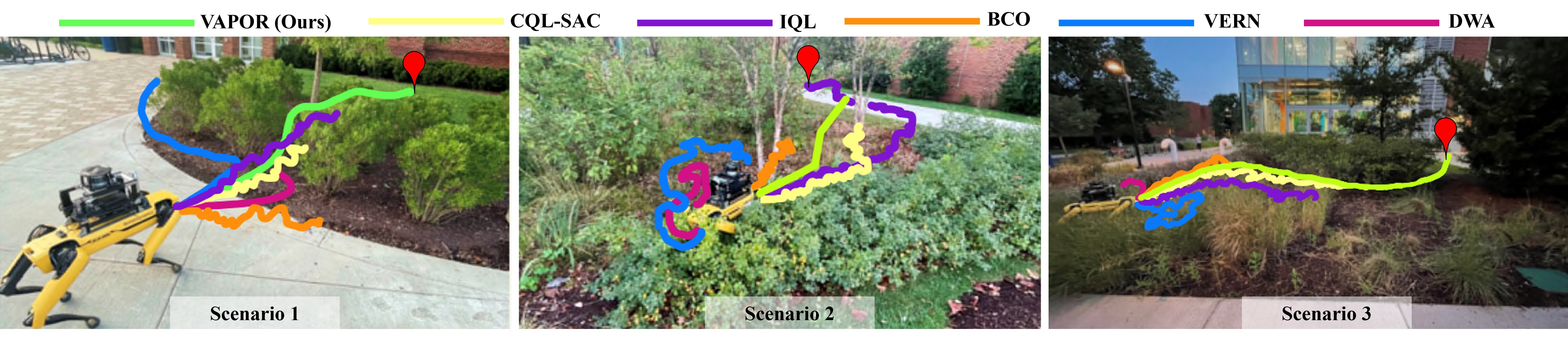}
    \caption{\small{Trajectories generated when navigating through complex outdoor vegetation using various comparison methods. The trajectories drawn from the robot's rear indicate that it has moved backward. \ours{} is able to use holonomic action in dense vegetation and vines (Scenarios 2 and 4) to reduce the risk of entrapment while others use angular velocities to reach the goal which results in instability and navigation failures. In the presence of narrow spaces in Scenarios 1 and 3, \ours{} uses non-holonomic actions to navigate through. }}
    \label{fig:comparison_trajs}
    \vspace{-15pt}
\end{figure*}

\begin{figure}[t]
      \centering
      \includegraphics[width=0.7\columnwidth,height=3.0cm]{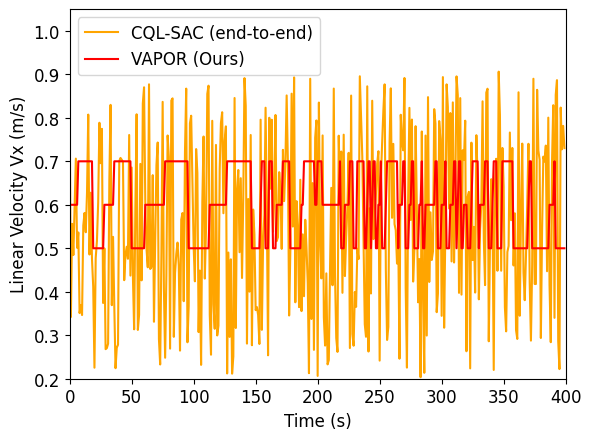}
      \caption {\small{Sample velocities $v_x$ generated by \ours{} and CQL-SAC (end-to-end) policy during a navigation task. The end-to-end policy actions demonstrate significantly high variance which indicates dynamically infeasible velocities to the robot leading to high vibrations and high motor current consumption. In contrast, \ours{} generates smooth, dynamically feasible actions. 
      }}
      \label{fig:vel_comparison}
      \vspace{-12pt}
\end{figure}

\subsection{Implementation}
Our CQL-SAC offline RL policy is implemented using PyTorch and our model is trained on a workstation with an Intel Xeon 3.6 GHz processor and an Nvidia Titan GPU. For real-time deployment and inference, we use the Spot robot from Boston Dynamics equipped with a  VLP16 Velodyne LiDAR, an onboard Intel NUC 11, which includes an Intel i7 CPU and an NVIDIA RTX 2060 GPU. 

\subsection{Comparison Methods and Evaluation Metrics}
We compare our method's navigation performance with three recent offline RL algorithms: CQL-SAC (our end-to-end policy) \cite{kumar2020CQL}, IQL \cite{kostrikov2021IQL}, BCO \cite{torabi2018BCO}, an autonomous imitation learning approach, VERN \cite{sathyamoorthy2023vern}, an outdoor vegetation navigation algorithm, and the Dynamic Window Approach (DWA) \cite{fox1997dynamic} a classical model-based navigation approach that uses 2D LiDAR scans. We train all the aforementioned offline RL comparison methods on our data set $\mathcal{D}$ using networks architectures similar to ours for fair comparison. We further perform two ablation studies: \ours{} w/o Proprioception; and \ours{} w/o attention to highlight the benefits of our approach. Our metrics for evaluation are:


\no \textbf{Success Rate} - The number of times the robot reached its goal while avoiding collisions with \textit{solid and dense vegetation} over the total number of attempts.

\no \textbf{Avg. Current Consumption} -  The average battery current consumption during a navigation task (i.e., $\sum_{traj} I_b$) in Amperes (A). 

\no \textbf{Normalized Traj. Length} - The robot’s trajectory length normalized using the straight-line distance to the goal for both successful and unsuccessful trajectories.


\subsection{Testing Scenarios}
We compare our method's navigation performance in the real-world outdoor test scenarios that are not included in the training data set. At least 10 trails are conducted in each scenario.

\begin{itemize}
\item \textbf{Scenario 1} - Contains narrow passages between shrubs, and trees in a mulch surface.

\item \textbf{Scenario 2} - Dense bushes that lead to entrapment, sparse grass, and trees.

\item \textbf{Scenario 3} - Thin grass, shrubs, and trees with narrow openings under low light conditions.

\item \textbf{Scenario 4} - Dense grass, fallen branches, vines, and trees.
\end{itemize}

\begin{table}
\resizebox{\columnwidth}{!}{%
\begin{tabular}{ |c |c |c |c |c |c |} 
\hline
\textbf{Metrics} & \textbf{Methods} & \multicolumn{1}{|p{1cm}|}{\centering \textbf{Scenario
1}} & \multicolumn{1}{|p{1cm}|}{\centering \textbf{Scenario
2}} & \multicolumn{1}{|p{1cm}|}{\centering \textbf{Scenario
3}} & \multicolumn{1}{|p{1cm}|}{\centering \textbf{Scenario
4}}\\ [0.5ex] 
\hline

\multirow{6}{*}{\rotatebox[origin=c]{0}{\makecell{\textbf{Success}\\\textbf{Rate (\%)}}}} 
 & DWA \cite{fox1997dynamic}  & 30 & 0 &  0 & 20   \\
 & VERN \cite{sathyamoorthy2023vern} & 60 &  \textbf{70} & 10 & 40 \\
 & BCO \cite{torabi2018BCO}  & 10 & 0 & 0 & 10 \\
 & IQL \cite{kostrikov2021IQL}  & 40 & 30 & 40 & 20 \\
 & CQL-SAC \cite{kumar2020CQL}& 50 & 60 & 50 & 50 \\
 &  \ours{} w/o Proprioception  & 50 & 40 & 50 & 30\\
 &  \ours{} w/o Attention  & 60 & 50 & 40 & 60 \\
 &  \ours{} (ours) & \textbf{80} &  \textbf{70} &  \textbf{60} &  \textbf{70}\\
\hline

\multirow{6}{*}{\rotatebox[origin=c]{0}{\makecell{\textbf{Avg. Current}\\\textbf{Consumption (A)}}}} 
 & DWA \cite{fox1997dynamic}  & 7.158 & 7.482 & 7.206 &  7.502  \\
 & VERN \cite{sathyamoorthy2023vern}& 6.937 & 7.457 & 6.993 & 7.423 \\
 & BCO \cite{torabi2018BCO}  & 6.681 & 7.153 & 6.937 & 7.391 \\
 & IQL \cite{kostrikov2021IQL} & 7.155 & 7.436 & 7.161 & 7.466 \\
 & CQL-SAC \cite{kumar2020CQL} & 7.192 & 7.301 & 7.099 & 7.487 \\
  &  \ours{} w/o Proprioception & 7.013 & 7.464 & 6.792 & 7.408\\
 &  \ours{} w/o Attention & 6.835 & 7.198 & \textbf{6.704} & \textbf{7.298}\\
 &  \ours{} (ours) & \textbf{6.599} & \textbf{7.147} & 6.735 & 7.319\\
\hline

\multirow{6}{*}{\rotatebox[origin=c]{0}{\makecell{\textbf{Norm. Traj.} \\\textbf{Length }}}}  
 & DWA \cite{fox1997dynamic}   & 1.327 & 1.655 & 0.428 & 1.421  \\
 & VERN \cite{sathyamoorthy2023vern} & 1.105 & 1.327 & 1.517 & 1.365 \\
 & BCO \cite{torabi2018BCO} & 0.425 & 1.398 & 0.422 & 0.643 \\
 & IQL \cite{kostrikov2021IQL} & 0.735 & 0.686 & 1.761 & 0.892 \\
 & CQL-SAC \cite{kumar2020CQL}& 0.897 & 1.245 & 1.453 & 1.277 \\
  &  \ours{} w/o Proprioception & 1.236 & 1.364 & 1.386 & 1.338\\
 &  \ours{} w/o Attention  & 1.125 & 1.223 & 1.374 & 1.294\\
 &  \ours{}(ours) & 1.065 & 1.238 & 1.289 & 1.256 \\
\hline


\end{tabular}
}
\caption{\small{Navigation performance of \ours{} compared to other methods on various evaluation metrics in four test scenarios that are not included in the data set. Please see \cite{vapor_arxiv} for more evaluations.}
}
\label{tab:comparison_table}
\end{table}

\begin{table}[t]
\centering
\resizebox{0.6\columnwidth}{!}{
\begin{tabular}{|c|c|} 
\hline
\textbf{Methods}\Tstrut \Tstrut & \textbf{Inference Time (ms)} \\ [0.5ex] 
\hline
VERN \cite{sathyamoorthy2023vern} & 84.612  \\
BCO \cite{torabi2018BCO} & 3.622 \\
IQL \cite{kostrikov2021IQL} & 3.951 \\

\ours{} w/o Attention & 8.820  \\

\ours{} (Ours) & 8.934  \\
\hline
\end{tabular}}

\caption{ \small{Inference time comparison between ours and other methods when executing in robot's onboard computer. IQL \cite{kostrikov2021IQL}, BCO \cite{torabi2018BCO}, and \ours{ w/o attention has the lowest inference time since they use the same network backbone. However, their navigation performance is significantly lower as shown in Table \ref{tab:comparison_table}. VERN \cite{sathyamoorthy2023vern} has the highest inference time due to computationally heavy backbones. In contrast, \ours{} has a lightweight network that can execute in real time while providing accurate predictions. }}}
\label{tab:inference_table}
\vspace{-18pt}
\end{table}

\subsection{Analysis and Comparison} \label{sec:analysis}

We evaluate our method's navigation performance qualitatively in the Fig. \ref{fig:comparison_trajs} and quantitatively in Table \ref{tab:comparison_table}. Scenario 4 is presented in the Fig. \ref{fig:cover-image}. We observe that \ours{} demonstrate the highest success rate compared to other methods in all four scenarios that include diverse and unseen vegetation.  Since the data set does not include expert demonstrations specifically collected with the behaviors imposed by reward functions, behavioral cloning with BCO \cite{torabi2018BCO} shows the lowest success rate due its attempt to imitate the data set trajectories without the knowledge of the rewards. In contrast, offline RL methods such as IQL and CQL-SAC attempt to perform the navigation tasks at a reasonable success rate. Eventhough VERN demonstrate the second best success rate in Scenarios 1 and 2, it performs poorly in low light conditions in Scenario 3 and trees covered with leaves in Scenario 4 due to the erroneous vegetation classification from its vision based system. DWA freezes in tall and dense vegetation in Scenarios 2 and 3 identifying such regions as obstacles from the 2D LiDAR scan.

\no \textbf{Benefits of Proprioception:} We observe that \ours{}'s performance in terms of success rate and current consumption degrades in dense vegetation without the proprioception state observations. Further, our planner uses proprioception to restrict the angular velocities during entrapment in scenarios 2 and 4 which leads to a higher success rate and low current consumption than \ours without prorioception. VERN and DWA leads to entrapment in scenario 2 and 4 due to lack of vegetation awareness from proprioception. Moreover, in stable conditions such as scenario 3, our planner uses angular velocities to move between the trees that create a narrow passage.

\no \textbf{Benefits of Attention:} Our method without attention demonstrate relatively low success rate, high power consumption and longer trajectory lengths particularly due to the lack of feature encoding capabilities between the cost map inputs than when spatial and channel attention are included. We observe that \ours{} without attention deviates from goal in some trails due to lack of spatial aware encoding from the goal cost map.

\no \textbf{End-to-end RL vs Ours:} We observe that end-to-end RL policies generates dynamically infeasible actions for the robot's motors though the actions reflects the behavior imposed by the rewards (See Fig. \ref{fig:vel_comparison}). This leads to jerky motion due to motor vibrations (see Fig. \ref{fig:comparison_trajs}) and high avg. current consumption. In contrast, \ours{}'s planner ensures that the actions are dynamically feasible which results in lower current consumption than all end-to-end RL models.

\no \textbf{Inference Time:} \ours{} has a lightweight network that can execute in real time ($\sim 112$Hz)  on the robot's onboard computer while providing accurate predictions as shown in Table \ref{tab:inference_table} and \ref{tab:comparison_table}. Vision based methods such as VERN \cite{sathyamoorthy2023vern} has a significantly lower inference time due to computationally heavy backbones. In contrast, \ours{} incorporate relatively lower dimensional state inputs that can represent $360^{\degree}$ view of the robot's vicinity and a light-weight network to obtain comparable or better navigation performance.


\section{Conclusions, Limitations and Future Work}
We present \ours{}, an offline-RL based method for legged robot navigation in outdoor vegetation. Our method uses randomly collected real world data to train a navigation policy that can reach local goals while avoiding dense and solid vegetation. Instead of end-to-end actions from the policy, its fully trained critic network is used to evaluate dynamically feasible actions generated by a planner. The planner is capable of adaptively switching between holonomic and non-holonomic action to minimize entrapment in unstructured vegetation. We deploy our method into a Boston Dynamics Spot robot and evaluate in real outdoor vegetation to demonstrate benefits.

Our method has a few limitations. Our planner cannot provide any theoretical guarantees on the behavior since the the state-action evaluations are obtained from a Q function trained on a data set. Even though our method generalizes well compared to vision based and supervised learning methods, large data sets are required for training. Further, our method cannot detect thin poles or string fences due to low resolution of the lidar and lack of scene awareness. 

\bibliographystyle{IEEEtran}
\bibliography{References}

\end{document}